\documentclass[letterpaper,10pt,conference]{ieeeconf}

\IEEEoverridecommandlockouts
\usepackage[utf8]{inputenc}
\usepackage[T1]{fontenc}
\usepackage{times}
\usepackage{graphicx}
\usepackage{amsmath,amssymb,bm}
\usepackage{booktabs,array,multirow}
\usepackage[table]{xcolor}
\usepackage{colortbl}
\usepackage{cite}
\usepackage{url}
\usepackage{placeins}
\usepackage{capt-of}


\graphicspath{{figures/}}
\newcommand{\method}{MAGNETAR}
\newcommand{\methodbase}{Multi-Arrival-Guided Neural Estimation over
Transmitter-Pose Alternatives from Radio}
\newcommand{\methodfull}{\methodbase}
\newcommand{\realNB}{\ensuremath{\mathrm{R}_{\mathrm{NB}}}}
\newcommand{\trainSR}{\ensuremath{\mathrm{S}{+}\mathrm{R}_{\mathrm{NB}}}}

\newcommand{\wrap}{\operatorname{wrap}}

\usepackage{dsfont}
\newcommand{\indicator}{\mathds{1}}

\newcommand{\bs}[1]{\boldsymbol{#1}}

\ifdefined
\fi
\ifdefined\pdfsuppressptexinfo
\fi

\title{\LARGE \bfseries MAGNETAR: Multipath-Guided Spatial Posteriors for
Transmitter Pose Inference in the Upper Mid-Band}

\author{Haozhe Lei$^{*}$, Ruibin Chen, Yuhan Jiang, Ali Rasteh,
  Aditya Dhananjay, and Sundeep Rangan%
  \thanks{\raggedright All authors are with NYU WIRELESS, Tandon School of
  Engineering, New York University, Brooklyn, NY 11201, USA. Emails:
  \texttt{\{hl4155,\allowbreak rc5018,\allowbreak yj3494,\allowbreak
  ar7655,\allowbreak avd263,\allowbreak srangan\}@nyu.edu}.
  $^{*}$Corresponding author: Haozhe Lei (\texttt{hl4155@nyu.edu}).}}

\IEEEaftertitletext{%
  \vspace{-12pt}%
  \begin{minipage}{\textwidth}
    \centering
    \includegraphics[width=\textwidth]{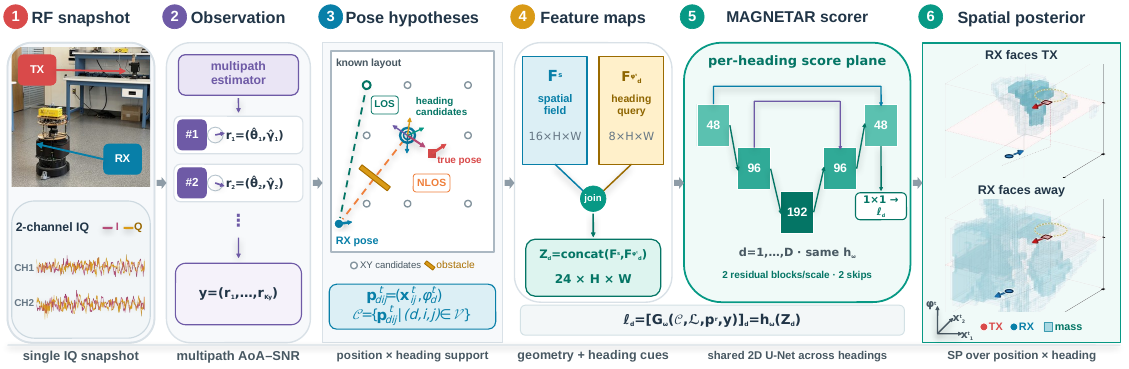}
    \captionof{figure}{Overview of \methodfull{} (\method{}) for transmitter
    (TX) pose inference. (1--2) A receiver (RX) records a two-channel
    radio-frequency (RF) in-phase/quadrature (IQ) snapshot; multipath estimation
    yields $\bs y$ as ranked angle of arrival (AoA) and signal-to-noise
    ratio (SNR) pairs. (3) Each candidate $\bs p^t_{dij}$ pairs XY position
    $\bs x^t_{ij}$ with heading $\phi^t_d$ (planar yaw); the known layout
    $\mathcal L$ and RX pose $\bs p^r$ determine candidate line-of-sight
    (LOS) or blocked non-line-of-sight (NLOS) visibility.
    (4--5) Spatial and heading features form $\bs Z_d$ (24 channels for
    $K=2$ retained arrivals), which a shared 2D U-Net maps to
    score plane $\boldsymbol\ell_d$. (6) One softmax over all XY--heading
    candidates yields the spatial posterior (SP). Horizontal axes give TX
    position; height gives heading, with the marked plane indicating the true
    heading. The two examples show how rotating the RX changes pose uncertainty.}
    \label{fig:global-overview}
  \end{minipage}
  \vspace{0.25\baselineskip}%
}

\begin{document}
\maketitle
\thispagestyle{empty}
\pagestyle{empty}
\raggedbottom

\begin{abstract}
Robots that localize a radio transmitter need more than a point
estimate: in cluttered rooms, one measurement is often consistent with several
transmitter locations and, because upper-mid-band antennas are directional, several
headings.  We present \method{}, which infers a joint posterior over planar transmitter
position and heading from a single asynchronous radio-frequency (RF) multipath
snapshot, represented by angle-of-arrival and signal-to-noise-ratio estimates,
given the room layout and receiver pose. Among our five neural scorers,
\method{} adopts a shared 2D U-Net conditioned on each candidate heading,
jointly normalizing scores over a discretized position--heading grid.
Training uses real-to-sim-calibrated 10~GHz simulations
and a small measured subset. Grid-based joint posteriors outperform
parametric ones on held-out simulations, the heading-conditioned scorer
transfers best to robotic measurements, and fusing joint posteriors improves
on fusing position-only marginals.
\end{abstract}


\section{Introduction}

Locating a radio transmitter (TX) from the signals it emits is a basic
capability for robots that must find a lost device, follow a tagged asset, or
navigate toward a wireless beacon~\cite{Ayyalasomayajula20DLoc,LeiICRA2024,Lei2025DTWIN}.
The upper mid-band (7--24~GHz)~\cite{3GPP38820} is attractive for this task:
wide bandwidths and compact arrays can resolve individual multipath
components~\cite{Bomfin_multiband}.  At these frequencies, however, practical
antennas are directional, so what a receiver (RX) observes depends not only on
where the TX is, but also on which way it faces.

Directivity couples position and heading in a way that one radio-frequency (RF) snapshot rarely
resolves.  In a cluttered room, blockage, reflections, and limited angular
resolution already make a single measurement consistent with several TX
locations.  Directivity adds a subtler ambiguity: a weak or missing arrival
from some direction may mean that no TX lies there, or that a TX lies there
but faces away from the RX.  The measurement therefore does not determine a
pose; it constrains a set of position--heading pairs whose shape depends on
the room, the RX pose, and the antenna patterns.  Fig.~\ref{fig:global-overview}
shows an example: the same TX yields a compact set when the RX faces it and a
broad one, spread over both position and heading, when the RX faces away.

This single fact has two consequences for what a localizer should output.
First, because the ambiguity is genuine rather than noise around a unique
answer, a point estimate discards what a robot needs in order to act: where
else the TX might be, and how confident to be before moving or searching.  A
\emph{posterior distribution} retains this information and can be fused
across observations and time~\cite{Arulampalam02,wang2025uncertainty,Lei25FullPosterior}.
Second, because the ambiguity is spread jointly over position and heading, the
posterior must be over the TX \emph{pose}: fusing position-only posteriors lets
views that imply incompatible TX headings reinforce the same wrong location,
whereas a joint posterior can require one heading across views.  This helps
when competing locations imply different headings: in one measured example it
keeps 81.1\% of the mass within 1~m of the TX versus 1.6\%, though across 1,000
simulated fusion sets the average gain is small
(Section~\ref{sec:results-fusion}).

We therefore seek the joint posterior of TX position and heading given the
observation, the RX pose, and the room layout.  It can be viewed as a 3D
heatmap, with two axes for position and one for heading, whose value is the
relative likelihood of each TX pose (Fig.~\ref{fig:global-overview}, panel~6).
Its shape changes with the room and the RX pose and is often multimodal or
elongated along heading, so simple parametric families fit it poorly.

We present \method{}, which learns this joint posterior, termed a spatial
posterior (SP), from a single asynchronous snapshot.  The RX uses only the
angle of arrival (AoA) and signal-to-noise ratio (SNR) of the strongest
multipath components, not their delays, so no TX--RX synchronization is
needed; a random $\pm3$~dB TX-power offset in training keeps power from
revealing range.  The SP is learned by scoring every candidate pose on a discretized grid and
normalizing jointly, which admits many scorers; we study five, from
per-candidate networks to 3D and 2D U-Nets.  \method{} treats each candidate
heading as a query to a shared 2D U-Net, so each heading hypothesis produces
its own spatial score map while all slices compete in one normalization.  Simulated tests use random TX headings, including a TX
facing away; robot measurements, with the TX facing the RX, validate
sim-to-real transfer.

\noindent\textbf{Contributions.}
\begin{itemize}
\item \emph{Joint pose posteriors for consistent fusion:} an explicit, jointly
normalized posterior over planar TX position and heading from one asynchronous
AoA--SNR snapshot, enabling shared-heading fusion before marginalization.
We develop five neural-posterior (NP) scorers (NP1--NP5), trained with
cross-entropy (CE) on a discretized pose grid, and adopt NP5 as MAGNETAR.
\item \emph{Mobile upper-mid-band platform and real-to-sim pipeline:} a
robot-mounted two-channel 10~GHz software-defined radio (SDR) RX with
light detection and ranging (LiDAR) mapping, a TX on a
motorized track and pan--tilt unit for controlled pose acquisition, and a
ray-tracing (RT) digital twin built from the simultaneous localization and mapping
(SLAM) map and chamber-measured antenna responses.  The measured dataset will
be released publicly.
\item \emph{Evaluation:} on 10,000 simulated tests with random TX headings,
grid-based joint posteriors outperform parametric Gaussian and mixture
posteriors, and the heading-conditioned U-Net (\method{}) is near-best; with
1,800 measured training records, it is the only U-Net scorer whose negative
log-likelihood (NLL) improves on 24,000 held-out measurements, where it ranks
first overall on the composite index. Shared-heading
fusion improves on fusing marginals.
\end{itemize}

\newcommand{\poseConeFigureWidth}{0.94\columnwidth}
\newcommand{\fusionFigureWidth}{0.92\columnwidth}
  \setlength{\textfloatsep}{5pt plus 2pt minus 1pt}
  \setlength{\dbltextfloatsep}{5pt plus 2pt minus 1pt}
  \setlength{\floatsep}{4pt plus 2pt minus 1pt}
  \setlength{\dblfloatsep}{4pt plus 2pt minus 1pt}
  \setlength{\intextsep}{4pt plus 2pt minus 1pt}
  \setlength{\abovedisplayskip}{3pt plus 1pt minus 1pt}
  \setlength{\belowdisplayskip}{3pt plus 1pt minus 1pt}
  \setlength{\abovedisplayshortskip}{0pt plus 1pt}
  \setlength{\belowdisplayshortskip}{3pt plus 1pt minus 1pt}
  \setlength{\jot}{1.5pt}
  \makeatletter
  \def\section{\@startsection{section}{1}{\z@}{0.8ex plus .4ex minus .2ex}%
  {0.3ex plus .2ex minus 0ex}{\normalfont\normalsize\centering\scshape}}
  \def\subsection{\@startsection{subsection}{2}{\z@}{0.8ex plus .4ex minus .2ex}%
  {0.3ex plus .2ex minus 0ex}{\normalfont\normalsize\itshape}}
  \def\@IEEEfigurecaptionsepspace{\vskip 4pt\relax}
  \let\magnetarOriginalMakeCaption\@makecaption
  \long\def\@makecaption#1#2{%
    \ifx\@captype\@IEEEtablestring
      {\centering\footnotesize #1\\{\scshape #2}\par}%
      \vskip 5pt\relax
    \else
      \magnetarOriginalMakeCaption{#1}{#2}%
    \fi}
  \makeatother
\noindent\textbf{Related Work.}
RF localization draws on environmental signal patterns
and propagation geometry. Fingerprinting links measurements to indoor
locations~\cite{HeChan16}. DLoc~\cite{Ayyalasomayajula20DLoc} maps multipath
measurements to room-frame images for robot navigation but regresses a
Gaussian-rendered target with L2 and L1 losses, giving a heatmap rather than a
normalized posterior; \method{} normalizes jointly over pose candidates and
trains with CE, a proper scoring rule. Resolved AoA,
angle of departure, and time of arrival enable joint position--orientation
estimation~\cite{Shahmansoori18,Nazari23Snapshot6D}. Multipath also supports
joint estimation of user state and propagation landmarks in snapshot radio
SLAM, under line-of-sight (LOS) and
non-line-of-sight (NLOS) propagation~\cite{Kaltiokallio25SnapshotSLAM}.

Spatial probability distributions retain alternative hypotheses for inference
and fusion. Position probability maps combine evidence across antennas and
access points~\cite{Gonultas22ProbabilityFusion}; temporal Bayesian methods
maintain beliefs across observations~\cite{Arulampalam02}, with learned AoA
uncertainty supporting access-point selection and fusion~\cite{zhang2024rloc}.
Candidate-based methods learn TX position posteriors from
single-path LOS AoA--SNR observations
(MC-CLE)~\cite{Lei25FullPosterior} or observed and ray-traced multipath
signatures (LOCUS-DT)~\cite{Lei26LOCUSDT}.
\method{} instead infers a joint TX-pose posterior from one snapshot.

Wireless digital twins provide scene-dependent radio training data
for localization~\cite{Morais24DTLocalization} and
navigation~\cite{Lei2025DTWIN}; physical priors also support zero-shot
navigation~\cite{LeiICRA2024,Li25PiPRL}.
Real-to-sim calibration connects propagation models to observed responses:
channel-response calibration addresses material and phase
mismatch~\cite{Ruah24RTCalibration}, while site-specific measurements support
location calibration and upper-mid-band RT validation~\cite{ying2026site}.
\method{} trains mostly on such a calibrated simulator
(Section~\ref{sec:hardware-data}).

\section{Problem Formulation}
\label{sec:method-problem}

Consider a TX at an unknown true planar pose
$\bs p^{t}_{0}=(\bs{x}^{t}_{0},\phi^{t}_{0})$, where
$\bs{x}^{t}_{0}\in\mathbb{R}^{2}$ is its position in a global coordinate
system and $\phi^{t}_{0}\in[-\pi,\pi)$ is its heading.  We use
$\bs p^{t}=(\bs{x}^{t},\phi^{t})$ for a generic TX pose.  To
infer the TX pose, the RX is given the room layout $\mathcal L$,
its pose $\bs p^r=(\bs x^r,\phi^r)$, and observations $\bs y$, where
$\bs x^r\in\mathbb R^2$ and $\phi^r\in[-\pi,\pi)$.

The TX broadcasts a known waveform over an asynchronous link; since path
delays are unused, no shared clock or phase reference is
needed~\cite{Nazari23Snapshot6D}.
The RX array is internally calibrated for AoA estimation. A fixed pipeline
$\mathcal E$ maps received in-phase/quadrature (IQ) samples
$\bs Y_{\mathrm{IQ}}$ to multipath estimates through path estimation,
rank selection, and SNR processing:
\begin{equation}
  \bs{y}=\mathcal E(\bs Y_{\mathrm{IQ}})
  =(\bs r_1,\ldots,\bs r_{K_y}),\quad
  \bs{r}_k=(\widehat{\theta}_k,\widehat{\gamma}_k),
  \label{eq:multipath-observation}
\end{equation}
where $K_y$ counts retained paths; $\widehat{\theta}_k$ and
$\widehat{\gamma}_k$ are their RX-local AoAs and processed SNRs.
AoA points from the RX toward the estimated source direction.
The pipeline can use any multipath estimator; our experiments give an example.

Given $\bs y$, we seek a posterior over a
discretized pose space rather than a single point estimate.  We partition the
modeled region into $H\times W$ spatial cells $\mathcal X^t_{ij}$ and the
heading circle into $D$ cells $\Phi^t_d$, with representatives
$\bs x^t_{ij}$ and $\phi^t_d$; spatial representatives retain row--column
order but may vary across supports.  A binary mask $m_{dij}\in\{0,1\}$
defines the valid index set $\mathcal V=\{(d,i,j):m_{dij}=1\}$, excluding
candidates coincident with the RX. The valid poses
$\bs p^t_{dij}=(\bs x^t_{ij},\phi^t_d)$ form the candidate set $\mathcal C$,
with $N_{\mathcal V}=|\mathcal V|$ candidates and a uniform prior.  The goal is
the $D\times H\times W$ SP
\begin{equation}
  p_{dij}=\Pr\!\big(\bs p^t_0\in\mathcal X^t_{ij}\times\Phi^t_d
  \,\big|\,\mathcal L,\bs p^r,\bs y\big),\quad (d,i,j)\in\mathcal V,
  \label{eq:tx-posterior}
\end{equation}
with $p_{dij}=0$ elsewhere.  Two axes index position and one indexes heading,
so the SP retains position--heading coupling.  Its position marginal is
$p_{XY}(i,j)=\sum_d p_{dij}$; summing over $(i,j)$ gives the heading marginal.

\section{Methodology}
\label{sec:methodology}

As illustrated in Fig.~\ref{fig:global-overview}, \methodfull{} (\method{})
learns a candidate score field whose joint normalization approximates the SP
in \eqref{eq:tx-posterior}.

\subsection{Score Function and Joint Candidate Posterior}
\label{sec:method-posterior}

\begin{figure}[!t]
  \centering
  \includegraphics[width=\columnwidth]{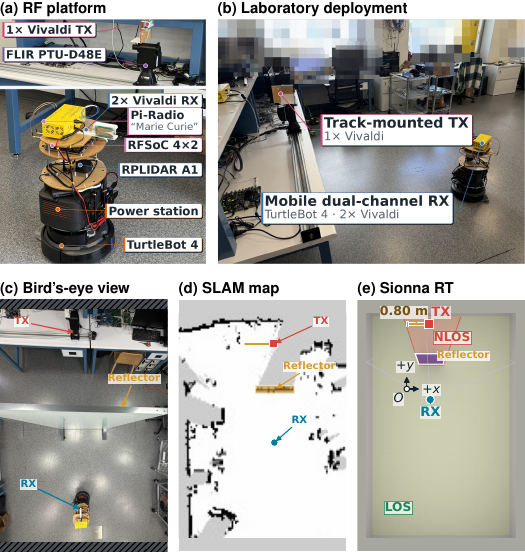}
  \caption{Hardware, site, and simplified room geometry.
  (a) RF acquisition uses one TX and two RX Vivaldi antennas, a Pi-Radio
  two-channel Frequency Range~3 (FR3) software-defined radio (SDR) kit
  (``Marie Curie''), and an RF system-on-chip (RFSoC) $4\times2$ board.
  SLAMTEC RPLIDAR A1 light detection and ranging (LiDAR) provides 2D SLAM measurements;
  the track and FLIR PTU-D48E pan--tilt unit set TX position and heading;
  the power station and TurtleBot~4 provide untethered RX power and mobility.
  (b,c) Laboratory and bird's-eye views. (d) SLAM map defining the room
  abstraction. (e) Aligned Sionna ray-tracing (RT) scene~\cite{sionna} with the middle
  reflector, TX track, RX pose, candidate LOS/NLOS regions, and reflected
  paths; axes define the measurement frame.}
  \label{fig:hardware-platform}
  \label{fig:site-middle-board}
\end{figure}

MAGNETAR learns $\boldsymbol\ell=G_{\boldsymbol\omega}
(\mathcal C,\mathcal L,\bs p^r,\bs y)
\in\mathbb R^{D\times H\times W}$ with parameters $\boldsymbol\omega$.
Each $\ell_{dij}$ scores a candidate against the conditioning information;
dependence on $\mathcal C$ permits pointwise and contextual scoring.
The estimate of \eqref{eq:tx-posterior} is
\mbox{$\widehat p_{\boldsymbol\omega,dij}=[\operatorname{softmax}_{\mathcal V}(\boldsymbol\ell)]_{dij}$}, jointly normalized
over valid candidates and zero elsewhere. Its XY marginal is
$\widehat p_{\boldsymbol\omega,XY}(i,j)=\sum_d\widehat p_{\boldsymbol\omega,dij}$.
Following candidate-based posterior learning~\cite{Lei25FullPosterior}, each
training support contains the truth position
$\bs x^t_{i^\star j^\star}=\bs x^t_0$.
Let $d_0,d_1$ be the circular neighbors of $\phi^t_0$, with linear
interpolation weights $w_0+w_1=1$.
For the joint candidate-mass loss
$\mathrm{CE}=-\sum_{\mathcal V}a_{dij}\log\widehat p_{\boldsymbol\omega,dij}$,
we use the uniform-referenced loss $L_{\mathrm U}=\mathrm{CE}-\log N_{\mathcal V}$:
\begin{equation}
  \begin{aligned}
    a_{dij}
      &=\indicator_{i=i^\star,j=j^\star}
        (w_0\indicator_{d=d_0}+w_1\indicator_{d=d_1}),\\
    L_{\mathrm U}
      &=-\sum_{\mathcal V}a_{dij}\ell_{dij}
      +\log\!\left(\frac{1}{N_{\mathcal V}}
      \sum_{\mathcal V}\exp(\ell_{dij})\right).
  \end{aligned}
  \label{eq:joint-cross-entropy}
\end{equation}
Here $\indicator_A$ equals 1 if $A$ holds and 0 otherwise, and
$\sum_{\mathcal V}$ sums over valid candidate indices. CE is NLL
for a one-hot target and its target-weighted extension
otherwise. We minimize the minibatch mean of $L_{\mathrm U}$; its
parameter-independent offset preserves CE gradients and minimizers and
sets the uniform posterior's loss to zero.

\subsection{Candidate Feature Representation}
\label{sec:method-input}

The key step is to map the multipath observation onto the candidate grid as
image-like feature fields, so that a standard 2D U-Net can score all
candidates with spatial context.  For each spatial candidate, let
$\beta_{ij}$ be the global RX-to-candidate bearing,
$\widetilde{\boldsymbol\delta}_{ij}$ and $\widetilde\rho_{ij}$ the normalized
RX-to-candidate displacement and log range, and $b_{\mathrm{LOS},ij}\in\{0,1\}$
indicate an unobstructed candidate--RX segment in $\mathcal L$; let
$\bs c(\alpha)=[\cos\alpha,\sin\alpha]$.  A heading-independent spatial field
$\bs F^s\in\mathbb R^{(8+4K)\times H\times W}$ stacks this geometry, the cell
spacing $(\Delta_x,\Delta_y)$, and, for each of the $K$ strongest arrivals, its
SNR and the mismatch $\bs c(\widehat\theta_k-(\beta_{ij}-\phi^r))$ between its
AoA and the candidate bearing.  A heading query
$\bs F^{\phi^t}\in\mathbb R^{(4+2K)\times D\times H\times W}$ compares each
candidate heading $\phi^t_d$ with the RX heading, the candidate-to-RX
direction, and each arrival's propagation direction.  Arrivals beyond the
$K_y\leq K$ available are zero-padded with a validity flag, and all features
use relative geometry and angle differences rather than absolute coordinates.
\subsection{TX-Heading-Conditioned Spatial Scorer}
\label{sec:method-magnetar}

\begin{samepage}
Our shared TX-heading-conditioned 2D U-Net~\cite{Ronneberger15UNet}
(Fig.~\ref{fig:global-overview}, stage~5) realizes $G_{\boldsymbol\omega}$
by concatenating the spatial field with each candidate heading query:
\begin{equation}
  \begin{aligned}
  \bs Z_d
    &=\operatorname{concat}
      (\bs{F}^{s},\bs{F}^{\phi^t}_{:,d,:,:})
      \in\mathbb R^{(12+6K)\times H\times W},\\
  \boldsymbol\ell_d
    &=h_{\boldsymbol\omega}(\bs Z_d)
      \in\mathbb{R}^{H\times W}.
  \end{aligned}
  \label{eq:heading-conditioned-scorer}
\end{equation}
\par\end{samepage}
Heading enters before the spatial encoder, allowing spatial context to depend
on $\phi^t_d$.

\subsection{Structured-Density Benchmarks}
\label{sec:structured-benchmarks}

We compare four RX-relative density families: a heading-wrapped Cartesian
Gaussian, Cartesian Gaussian-mixture models (GMMs)~\cite{Bishop94} with two
or three components, and an area-polar Gaussian with wrapped bearing and
heading. All components use full covariance matrices.
Cartesian cores use RX-local XY and $\wrap(\phi^t-\phi^r)$, where $\wrap$
maps angles to $[-\pi,\pi)$.
Area-polar cores use
$(\rho^2/(2L_{\mathrm{ref}}^2),\wrap(\beta-\phi^r),\wrap(\phi^t-\phi^r))$,
with range $\rho$, global bearing $\beta$, and $L_{\mathrm{ref}}=1$~m.
Their pose-space Jacobian factors are $1$ and $L_{\mathrm{ref}}^{-2}$,
respectively. Periodic image sums use indices $-2,\ldots,2$ per wrapped
coordinate.

Let $f_{\mathrm{core}}(\bs p^t\mid\bs p^r,\bs y)$ denote a core's
pose-space density, suppressing its dependence on candidate spacing.
Its score on the common candidate support is
\begin{equation}
  \begin{aligned}
  \ell^{\mathrm{str}}_{dij}
  &=\log f_{\mathrm{core}}\!\left(
    \bs p^t_{dij}\mid\bs p^r,\bs y\right)+\log V_{dij}\\
  &\quad+b_{\mathrm{LOS},ij}
    \eta_{\mathrm{LOS}}(\widetilde{\boldsymbol\delta}_{ij},\widetilde\rho_{ij},
    \bs c(\beta_{ij}-\phi^r)),
  \end{aligned}
  \label{eq:structured-density-score}
\end{equation}
where $\eta_{\mathrm{LOS}}(\cdot)$ is a bounded learned LOS adjustment.  The term
$\log V_{dij}$, with nominal cell measure $V_{dij}=\Delta_x\Delta_y(2\pi/D)$,
converts the density to candidate mass. The same
mask, joint softmax, and training objective in
\eqref{eq:joint-cross-entropy} apply.

\section{Experimental Design}
\label{sec:experimental-design}

Sim and Real denote simulated and measured observations.
S denotes Sim-only training; \trainSR{} adds the no-board Real training
subset, denoted \realNB{} (NB: no board).
Both regimes are evaluated on separate Sim and Real test sets
(Table~\ref{tab:data-inventory}).

\subsection{Architectural Variants and Benchmarks}
\label{sec:model-families}

\begin{table}[!t]
  \centering
  \caption{Neural scoring variants and parametric benchmarks.}
  \label{tab:model-families}
  \footnotesize
  \renewcommand{\arraystretch}{1.00}
  \begin{tabular*}{\columnwidth}{@{\extracolsep{\fill}}ll@{}}
    \hline
    Code & Model \\
    \hline
    NP1 & Independent candidate MLP \\
    NP2 & Candidate-to-path attention \\
    NP3 & 3D residual U-Net \\
    NP4 & 2D residual U-Net with pointwise heading queries \\
    NP5 & MAGNETAR: heading-conditioned 2D U-Net \\
    SM1 / SA1 & Wrapped Cartesian Gaussian \\
    SM2 / SA2 & Wrapped area-polar Gaussian \\
    SM3 / SA3 & Two-component Cartesian GMM \\
    SM4 / SA4 & Three-component Cartesian GMM \\
    \hline
  \end{tabular*}
  \vspace{2pt}

  \parbox{\columnwidth}{\footnotesize
  NP: our neural-posterior variants. SM and SA use multilayer perceptron (MLP)
  and attention conditioners, respectively.}
\end{table}

All designs in Table~\ref{tab:model-families} share $(\mathcal L,\bs p^r,\bs y)$
and $\mathcal C$. We develop NP1--NP5 as joint-SP scoring variants;
SM/SA are parametric benchmarks.
With $K=2$, NP1--NP5 use the features in
$\bs Z_d\in\mathbb R^{24\times H\times W}$ from
\eqref{eq:heading-conditioned-scorer}.
NP2 independently applies two-head attention from each candidate to the observed paths.
NP3 uses circular heading padding; NP4 applies heading queries pointwise
after a heading-independent spatial encoder.

We adopt NP5 as MAGNETAR, with heading-conditioned spatial encoding
(Section~\ref{sec:method-magnetar}).  Its shared 2D residual U-Net uses two
downsampling levels, widths 48/96/192, two residual blocks per level with group
normalization and sigmoid linear unit (SiLU) activations, skip-connected
nearest-neighbor decoding, and a final $1\times1$ logit projection.

SM and SA use the density cores in Section~\ref{sec:structured-benchmarks}
and the bounded LOS adjustment in \eqref{eq:structured-density-score};
SA uses validity-masked two-head path attention.

\begin{table}[!t]
  \centering
  \caption{Dataset sizes (observations). Test data are held out.}
  \label{tab:data-inventory}
  \footnotesize
  \renewcommand{\arraystretch}{1.04}
  \begin{tabular*}{\columnwidth}{@{\extracolsep{\fill}}llrrr@{}}
    \toprule
    Use & Source & No board & Board & Total \\
    \midrule
    \multirow{2}{*}{Train} & Sim  & 32,000 & 48,000 & 80,000 \\
                          & Real &  1,800 &      0 &  1,800 \\
    \midrule[0.3pt]
    Validation            & Sim  &  8,000 & 12,000 & 20,000 \\
    \midrule[0.3pt]
    \multirow{2}{*}{Test}  & Sim  &  4,000 &  6,000 & 10,000 \\
                          & Real &  5,580 & 18,420 & 24,000 \\
    \bottomrule
  \end{tabular*}
\end{table}

\subsection{Measurement and Simulation Setup}
\label{sec:hardware-data}
\label{sec:sre-estimator}

Fig.~\ref{fig:hardware-platform}(a,b) shows the measurement platform and LiDAR.
Sim and Real use an ordered two-element RX array with nominal 26~mm spacing
and a 10~GHz carrier in the 7--24~GHz range studied by
the 3rd Generation Partnership Project (3GPP) for New Radio~\cite{3GPP38820},
commonly termed Frequency Range~3 (FR3) in upper-mid-band research.
Both use 491.52~MHz processing bandwidth over 512 frequency bins.
The measurement frame in Fig.~\ref{fig:site-middle-board}(e) has a fixed
laboratory reference origin; device positions are rotating-mount centers.
The TX pan--tilt mount follows a 0.8~m linear track from $(0.18,2.28)$~m
to $(0.98,2.28)$~m.
Hardware limits TX rotation to the $180^\circ$ sector facing the RX sampling region;
candidate headings cover $360^\circ$ with $D=18$ bins.
Laboratory space limits nominal Real RX sampling to $x\in[0,2]$~m and
$y\in[-2,0]$~m, with full RX-yaw coverage.

We process Sim and Real IQ with the same sparse-refinement estimator
(SRE)~\cite{Bomfin_multiband,Lei26LOCUSDT} and retain the two strongest
components ($K=2$) by raw SNR.  Their processed SNRs and RX-local AoAs form
$\bs y$ in \eqref{eq:multipath-observation}; missing ranks follow the validity convention in
Section~\ref{sec:method-input}.

Chamber measurements provide inter-channel phase calibration and phase-free
TX/RX angular responses for the antenna adapters.  The RF chain has no
absolute amplitude calibration, so IQ magnitudes represent relative rather
than absolute received power.  For Sim, measured responses are
applied to each ray-traced path before noise and SRE; Real IQ already contains
the antenna and RF-chain responses.

\begin{figure}[!t]
  \centering
  \includegraphics[width=\columnwidth]{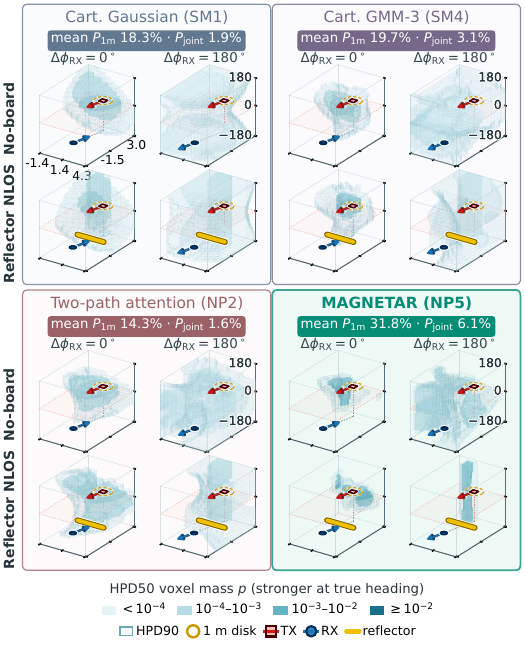}
  \caption{Real SP volumes for four representative \trainSR{} designs
  (Table~\ref{tab:model-families}), using one common seed and
  $18\times33\times33$ support. Rows show no-board reference and held-out
  reflector NLOS; columns show RX offsets $\Delta\phi_{\mathrm{RX}}=0^\circ,180^\circ$.
  Height is truth-relative TX heading. Dark 50\% highest posterior density
  (HPD) voxels use shared mass bins; pale envelopes show 90\% HPD.
  Emphasis marks the nearest truth-heading bin. Badges average four cases:
  $P_{1\mathrm m}$ is mass within 1~m of truth XY over all headings;
  $P_{\mathrm{joint}}$ also requires the nearest truth-heading bin.}
  \label{fig:joint-pose-volume}
\end{figure}

Real-to-sim preparation combines this calibration
with a simplified six-surface SLAM-derived
room (Fig.~\ref{fig:site-middle-board}(d)) in Sionna RT~\cite{sionna}, a
radio-propagation simulator. Nominal room bounds are
$x\in[-1.403,4.270]$~m, $y\in[-6.060,3.000]$~m, and $z\in[0,3.050]$~m.
The reflector measures
$1.015\times0.033\times1.715$~m (width $\times$ thickness $\times$ height).
Left, middle, and right placements are centered at $(x,1,1.0025)$~m,
with $x=0.2275,1.0225,1.7625$, respectively, and width aligned with the $x$-axis.
The room model retains only outer surfaces.
Measured wall and board responses guide reference-scene material choices.
The reflector uses Sionna's glass model;
RT includes LOS and first-order specular reflections, with transmission disabled.
Calibrated paths are summed into $512\times2$ frequency-domain observations.
Sim adds independent unit-variance complex Gaussian noise after calibrated
scaling and energy-preserving RF-chain perturbations, with $-19.14$~dB
input-power and $+18.89$~dB output-SNR adjustments. A random $\pm3$~dB
TX-power offset per Sim observation keeps received power from determining range. Real uses inactive-bin noise normalization.

\begin{figure}[!t]
  \centering
  \includegraphics[width=0.90\columnwidth]{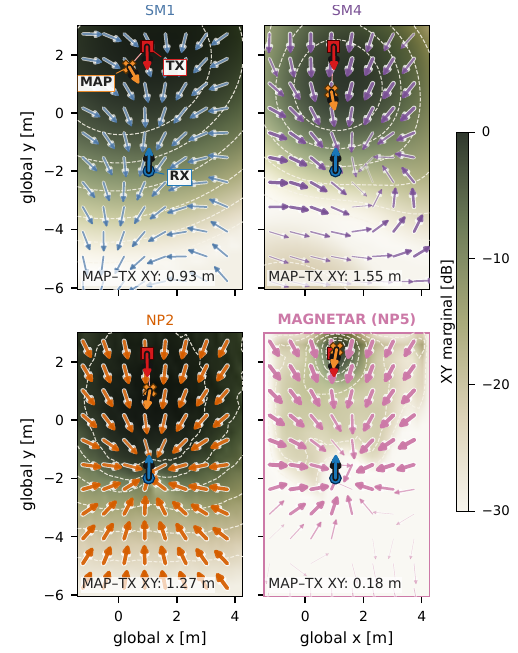}
  \caption{XY marginal and conditional-heading field for the four \trainSR{} designs
  in Fig.~\ref{fig:joint-pose-volume}, using one held-out no-board Real record
  and the same seed. Background is $\widehat p_{\boldsymbol\omega,XY}$ in
  peak-relative dB. Arrows show mean TX heading; opacity and width show
  circular concentration. Orange marks the joint posterior mode; red/blue
  mark TX/RX. Labels give its XY error. Pale contours:
  $-25,-20,-15,-10,-6,-3$~dB.}
  \label{fig:pose-heading-field}
\end{figure}

\subsection{Sim and Real Data Distributions}
\label{sec:sim-real-data}

Table~\ref{tab:data-inventory} lists dataset sizes by split and board condition.
Sim training uses one of three (wall, ceiling, floor) assignments:
(concrete, concrete, marble), (brick, concrete, concrete), or
(plasterboard, ceiling board, wood). Room planar scale varies by up to 5\%, aspect
ratio and height by 2\%, and wall translation by 0.2~m per planar axis.
The Sim reflector varies across admissible link-relative
regions, with height $1.27$~m, width $0.75$--$1.25$~m, center height
$1.08+\{-0.2,0,0.2\}$~m, and yaw jitter bounded by $\pm10^\circ$.
TX/RX positions are sampled continuously within each room, and their
headings independently and uniformly over $[0,2\pi)$.
The single-observation Sim test set uses disjoint realizations of the
same rectangular-room family, translations of $0.20$--$0.35$~m per axis
in either direction, and held-out material assignments
(wood, ceiling board, concrete) or (glass, concrete, marble).
The detailed measured room is not used as a training geometry.
For each realization, the room bounds define the candidate support, while
the optional reflector-board footprint defines $b_{\mathrm{LOS},ij}$
(Section~\ref{sec:method-input}).

Both Real acquisition types below include no-board data and the three
reference board placements.

\begingroup\clubpenalty=10000\widowpenalty=10000
\emph{Full-rotation sweeps.} With TX fixed at $(0.98,2.28)$~m, RX heading
commands span $0^\circ$--$359^\circ$ in $1^\circ$ steps at seven nominal sites (metres):
$(1,0)$, $(x,-1)$ for $x=0.5,1,1.5$, and $(x,-2)$ for $x=0,1,2$.
\par\endgroup

\emph{Track scans.} The 25 nominal RX sites form a 0.5~m grid over
$x\in[0,2]$~m and $y\in[-2,0]$~m. Each site uses three or six RX heading
commands without the board and six with it, drawn independently and
uniformly from $[0^\circ,360^\circ)$. Each heading block contains ten
TX-track samples.

\begingroup\widowpenalty=10000
Processing and evaluation use recorded device positions and RX headings;
TX heading follows the TX-to-RX bearing under the RX-facing acquisition assumption. The registered
board condition determines $b_{\mathrm{LOS},ij}$ during evaluation.
\par\endgroup

S trains on Sim only; \trainSR{} adds 4,096 no-board Real exposures per run
(0.25\% of Sim) at equal per-sample loss weight, keeping the Sim budget
unchanged. The 1,800 Real source records are cycled with common TX/RX
translations preserving relative geometry and headings, and are disjoint
from all test records.

\begin{figure*}[!t]
  \centering
  \includegraphics[width=0.92\textwidth]{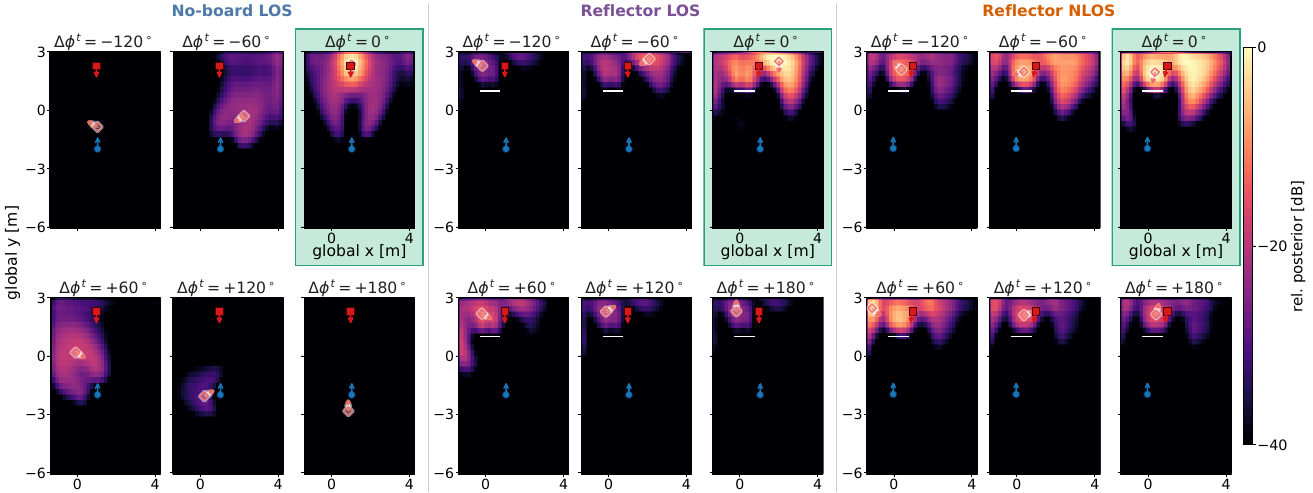}
  \caption{Fixed-heading SP slices of \trainSR{} MAGNETAR (NP5) for three held-out
  Real records, recomputed on truth-aligned $20^\circ$ heading supports.
  Blocks show no-board LOS, reflector LOS, and reflector NLOS.
  Panels are labeled by truth-relative heading $\Delta\phi^t$
  ($0^\circ$: truth; $+180^\circ$: antipodal).
  Slices retain joint normalization and share each record's
  full-volume peak dB reference. Diamonds mark dominant-mode centroids;
  their arrows give each panel's candidate heading.}
  \label{fig:heading-slice-sweep}
\end{figure*}

\subsection{Training Protocol}
\label{sec:model-training}

All models use Adam and the uniform-referenced loss
$L_{\mathrm U}$ in \eqref{eq:joint-cross-entropy}. The peak learning rate (LR)
is $5\times10^{-4}$, except $1.5\times10^{-4}$ for NP3 and
$2.25\times10^{-4}$ for NP4/NP5. Each run has 1,638,400 Sim exposures
in four equal stages with $33^2,25^2,33^2,41^2$ spatial supports and
$D=18$ headings. One LR schedule spans all stages: 2\% linear warmup
then cosine decay to zero. Training uses the target in
\eqref{eq:joint-cross-entropy}, independently redrawing nontruth axis
representatives within their uniform strata. We select among the four
stage-end models by density NLL (mass NLL plus $\log V_{dij}$) averaged over
the 20,000 Sim validation records and three distinct support sizes.

Quantitative Sim and Real evaluation uses fixed $18\times33\times33$ supports
spanning each room, without inserting truth. Bilinear XY and circular-linear heading
interpolation project truth onto at most eight voxels, forming the
normalized target scored by raw $\mathrm{CE}$
(Section~\ref{sec:method-posterior}).  S and \trainSR{} use the same five random seeds.
Each run uses one NVIDIA H100 NVL graphics processing unit (GPU); the 130 runs
require approximately 65 GPU-hours in total.

\section{Experimental Results}
\label{sec:results}

\setcounter{topnumber}{1}

\begin{figure}[t]
  \centering
  \includegraphics[width=\poseConeFigureWidth]{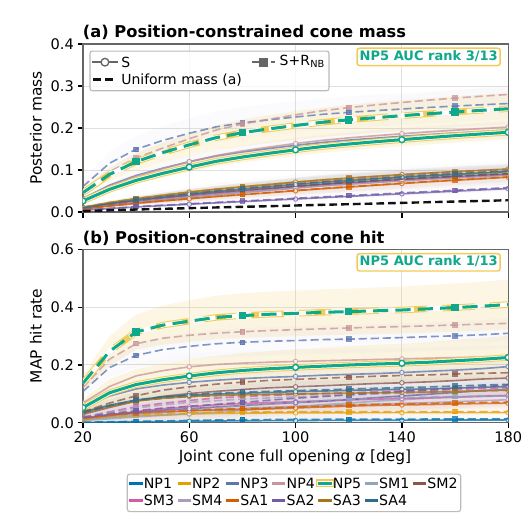}
  \caption{Held-out Real position-constrained cone mass (a) and joint-MAP
  hit rate (b). Both require XY error $\leq1$~m and bearing and TX-heading
  errors $\leq\alpha/2$; (b) tests the global joint MAP. For 13 designs
  under S and \trainSR{}, curves and bands show domain-, protocol-,
  and site-balanced mean $\pm$ sample standard deviation over five seeds.
  The uniform reference applies only to mass. Codes follow
  Table~\ref{tab:model-families}.}
  \label{fig:pose-cone-benchmark}
\end{figure}

Figs.~\ref{fig:joint-pose-volume}--\ref{fig:fusion-orders} use measured Real
observations for controlled sim-to-real validation; qualitative examples
use \trainSR{} models with one common seed. Table~\ref{tab:pose-global-probabilistic-five-seed}
compares both Sim and Real test sets across five seeds.

\subsection{Joint Position--Heading Structure}
\label{sec:results-joint-output}

In Fig.~\ref{fig:joint-pose-volume},
$\Delta\phi^t=\wrap(\phi^t_d-\phi^t_0)$ is truth-relative candidate heading,
and $\Delta\phi_{\mathrm{RX}}$ is RX-heading offset from the sweep reference.
SM1/SM4 retain broad volumes and separated lobes, respectively; NP2 also
remains broad. MAGNETAR forms a compact no-board mode at
$\Delta\phi_{\mathrm{RX}}=0^\circ$, with broader alternatives at
$\Delta\phi_{\mathrm{RX}}=180^\circ$. Under reflector NLOS, XY remains localized
across headings: the SP preserves position information despite orientation ambiguity.

\begingroup
\leavevmode\clubpenalty=10000\widowpenalty=10000
Figure~\ref{fig:pose-heading-field} summarizes
$\widehat p_{\boldsymbol\omega}(d\mid i,j)\allowbreak=\nobreak
\widehat p_{\boldsymbol\omega,dij}/\widehat p_{\boldsymbol\omega,XY}(i,j)$
by circular mean and concentration where XY mass is nonzero.
NP2's coherent RX-facing headings accompany a displaced joint maximum a
posteriori (MAP) position; NP5 aligns high XY mass and heading at the TX.
This contrast supports heading-conditioned spatial context beyond
candidate-wise path matching.\par
\endgroup

Figure~\ref{fig:heading-slice-sweep} exposes alternatives hidden by the mean
through jointly normalized fixed-heading slices. For no-board LOS,
mismatched headings move or suppress the truth-matched TX mode. With a
reflector, separated modes exchange strength in LOS, while NLOS support
persists across headings. The SP thus retains alternative locations and
their associated headings.

\providecommand{\meanstd}[2]{\ensuremath{#1\,{\scriptstyle\pm}\,#2}}
\begin{table*}[!t]
  \centering
  \caption{Held-out Sim/Real performance, execution time, and Real-based index.}
  \label{tab:pose-global-probabilistic-five-seed}
  \label{tab:execution-posterior-index}
  \footnotesize
  \newcommand{\tablebest}[1]{\textcolor{blue!70!black}{#1}}
  \newcommand{\tablesecond}[1]{\textcolor{red!75!black}{#1}}
  \newcommand{\tablethird}[1]{\textcolor{green!45!black}{#1}}
  \setlength{\tabcolsep}{2pt}
  \setlength{\aboverulesep}{0.3ex}
  \setlength{\belowrulesep}{0.45ex}
  \renewcommand{\arraystretch}{1.00}
  \begin{tabular*}{\textwidth}{@{\extracolsep{\fill}}>{\columncolor{white}[0pt][\tabcolsep]}llrrrrrrr>{\columncolor{white}[\tabcolsep][0pt]}r@{}}
    \multicolumn{10}{@{}l}{\textbf{(a) Posterior performance}} \\[1pt]
    \toprule
    & & \multicolumn{2}{c}{Sim} & \multicolumn{6}{c}{Real} \\
    \cmidrule(lr){3-4}\cmidrule(l){5-10}
    Model & Train & \shortstack{NLL$\downarrow$\\[1pt]{[nat]}} &
    \shortstack{$V@40\%\downarrow$\\[1pt]{[\%]}} & \shortstack{NLL$\downarrow$\\[1pt]{[nat]}} &
    \shortstack{HPD gap$\downarrow$\\[1pt]{[pp]}} & \shortstack{$V@40\%\downarrow$\\[1pt]{[\%]}} &
    \shortstack{MAP XY$\downarrow$\\[1pt]{[m]}} & \shortstack{MAP yaw$\downarrow$\\[1pt]{[$^\circ$]}} &
    \shortstack{Joint hit$\uparrow$\\[1pt]{[\%]}} \\
    \midrule
NP1 & S & \meanstd{9.377}{0.012} & \meanstd{16.8}{0.5} & \meanstd{8.875}{0.069} & \meanstd{8.1}{2.0} & \meanstd{7.9}{0.8} & \meanstd{2.75}{0.18} & \meanstd{59.7}{2.3} & \meanstd{0.3}{0.1} \\
NP1 & \trainSR{} & \meanstd{9.376}{0.013} & \meanstd{16.7}{0.6} & \meanstd{8.840}{0.069} & \meanstd{9.0}{2.4} & \meanstd{7.5}{1.0} & \meanstd{2.72}{0.19} & \meanstd{59.7}{2.7} & \meanstd{0.5}{0.3} \\
    \addlinespace[0.6pt]
NP2 & S & \meanstd{9.289}{0.013} & \meanstd{14.9}{0.3} & \meanstd{8.691}{0.058} & \meanstd{5.9}{1.0} & \meanstd{7.8}{1.6} & \meanstd{2.64}{0.20} & \meanstd{61.8}{1.8} & \meanstd{2.4}{2.0} \\
NP2 & \trainSR{} & \meanstd{9.290}{0.013} & \meanstd{14.9}{0.3} & \meanstd{8.668}{0.058} & \meanstd{6.4}{1.4} & \meanstd{7.4}{1.6} & \meanstd{2.61}{0.21} & \meanstd{60.9}{1.9} & \meanstd{2.7}{2.1} \\
    \addlinespace[0.6pt]
NP3 & S & \tablesecond{\meanstd{8.585}{0.014}} & \tablesecond{\meanstd{10.8}{0.1}} & \meanstd{8.500}{0.236} & \meanstd{21.3}{6.1} & \meanstd{6.2}{1.0} & \meanstd{2.25}{0.13} & \meanstd{59.2}{2.8} & \meanstd{11.7}{2.8} \\
NP3 & \trainSR{} & \meanstd{8.602}{0.033} & \meanstd{11.0}{0.2} & \meanstd{9.182}{0.840} & \meanstd{21.5}{9.1} & \tablethird{\meanstd{5.2}{1.1}} & \tablesecond{\meanstd{1.69}{0.05}} & \tablesecond{\meanstd{43.9}{1.2}} & \tablethird{\meanstd{23.5}{4.2}} \\
    \addlinespace[0.6pt]
NP4 & S & \tablebest{\meanstd{8.570}{0.009}} & \tablebest{\meanstd{10.7}{0.2}} & \tablethird{\meanstd{8.413}{0.134}} & \meanstd{16.9}{4.2} & \meanstd{5.8}{0.8} & \tablethird{\meanstd{2.21}{0.15}} & \meanstd{58.6}{5.8} & \meanstd{16.3}{4.5} \\
NP4 & \trainSR{} & \tablethird{\meanstd{8.598}{0.065}} & \tablethird{\meanstd{10.9}{0.4}} & \meanstd{9.277}{1.266} & \meanstd{19.3}{5.3} & \tablesecond{\meanstd{4.9}{0.4}} & \tablebest{\meanstd{1.62}{0.03}} & \tablebest{\meanstd{41.4}{2.1}} & \tablesecond{\meanstd{27.5}{2.7}} \\
    \addlinespace[0.6pt]
\rowcolor{yellow!12} \textbf{NP5$^\star$} & S & \meanstd{8.628}{0.031} & \tablesecond{\meanstd{10.8}{0.4}} & \tablesecond{\meanstd{8.202}{0.217}} & \meanstd{12.2}{7.6} & \tablethird{\meanstd{5.2}{1.2}} & \meanstd{2.31}{0.09} & \meanstd{60.7}{5.3} & \meanstd{13.3}{5.4} \\
\rowcolor{yellow!12} \textbf{NP5$^\star$} & \trainSR{} & \meanstd{8.641}{0.014} & \tablethird{\meanstd{10.9}{0.4}} & \tablebest{\meanstd{7.774}{0.232}} & \meanstd{6.9}{6.1} & \tablebest{\meanstd{3.4}{1.0}} & \tablesecond{\meanstd{1.69}{0.20}} & \tablethird{\meanstd{47.2}{4.9}} & \tablebest{\meanstd{31.5}{6.0}} \\
    \midrule
SM1 & S & \meanstd{9.542}{0.015} & \meanstd{20.1}{0.6} & \meanstd{9.213}{0.131} & \meanstd{7.5}{2.8} & \meanstd{16.3}{5.0} & \meanstd{2.35}{0.22} & \meanstd{76.7}{3.9} & \meanstd{2.0}{1.0} \\
SM1 & \trainSR{} & \meanstd{9.541}{0.013} & \meanstd{20.1}{0.5} & \meanstd{9.071}{0.077} & \meanstd{4.5}{2.6} & \meanstd{13.3}{3.7} & \meanstd{2.31}{0.19} & \meanstd{74.1}{4.7} & \meanstd{3.1}{1.4} \\
    \addlinespace[0.6pt]
SM2 & S & \meanstd{9.536}{0.039} & \meanstd{21.6}{2.2} & \meanstd{9.014}{0.339} & \meanstd{9.7}{10.0} & \meanstd{10.3}{6.5} & \meanstd{2.91}{0.31} & \meanstd{73.9}{14.0} & \meanstd{7.4}{4.2} \\
SM2 & \trainSR{} & \meanstd{9.536}{0.039} & \meanstd{21.7}{2.2} & \meanstd{8.997}{0.338} & \meanstd{11.0}{10.1} & \meanstd{10.0}{6.1} & \meanstd{2.90}{0.30} & \meanstd{68.8}{4.8} & \meanstd{9.3}{1.3} \\
    \addlinespace[0.6pt]
SM3 & S & \meanstd{9.495}{0.011} & \meanstd{18.8}{0.6} & \meanstd{9.230}{0.318} & \meanstd{6.8}{5.5} & \meanstd{15.4}{4.7} & \meanstd{2.69}{0.09} & \meanstd{64.0}{4.3} & \meanstd{3.9}{2.5} \\
SM3 & \trainSR{} & \meanstd{9.493}{0.013} & \meanstd{18.7}{0.7} & \meanstd{8.996}{0.206} & \tablesecond{\meanstd{3.8}{2.0}} & \meanstd{12.0}{3.0} & \meanstd{2.64}{0.08} & \meanstd{63.3}{4.4} & \meanstd{5.6}{2.6} \\
    \addlinespace[0.6pt]
SM4 & S & \meanstd{9.470}{0.037} & \meanstd{18.4}{0.8} & \meanstd{9.248}{0.368} & \meanstd{9.9}{5.8} & \meanstd{17.8}{6.6} & \meanstd{2.64}{0.08} & \meanstd{73.6}{12.8} & \meanstd{2.8}{2.7} \\
SM4 & \trainSR{} & \meanstd{9.460}{0.021} & \meanstd{18.3}{0.5} & \meanstd{8.990}{0.259} & \meanstd{5.3}{4.3} & \meanstd{12.9}{4.1} & \meanstd{2.63}{0.06} & \meanstd{67.8}{10.7} & \meanstd{5.2}{3.6} \\
    \midrule
SA1 & S & \meanstd{9.549}{0.022} & \meanstd{20.2}{0.8} & \meanstd{9.219}{0.147} & \meanstd{6.2}{4.3} & \meanstd{12.6}{3.3} & \meanstd{2.42}{0.30} & \meanstd{83.8}{10.7} & \meanstd{3.3}{3.0} \\
SA1 & \trainSR{} & \meanstd{9.550}{0.022} & \meanstd{20.2}{0.7} & \meanstd{9.201}{0.156} & \meanstd{5.8}{4.6} & \meanstd{12.0}{2.9} & \meanstd{2.41}{0.31} & \meanstd{82.2}{9.8} & \meanstd{3.4}{3.1} \\
    \addlinespace[0.6pt]
SA2 & S & \meanstd{9.589}{0.008} & \meanstd{23.7}{0.4} & \meanstd{9.549}{0.083} & \meanstd{13.1}{6.5} & \meanstd{14.2}{2.0} & \meanstd{2.59}{0.20} & \meanstd{90.6}{6.3} & \meanstd{3.5}{2.9} \\
SA2 & \trainSR{} & \meanstd{9.587}{0.012} & \meanstd{23.7}{0.4} & \meanstd{9.504}{0.165} & \meanstd{15.7}{6.6} & \meanstd{12.6}{5.1} & \meanstd{2.50}{0.27} & \meanstd{86.4}{18.1} & \meanstd{4.1}{4.3} \\
    \addlinespace[0.6pt]
SA3 & S & \meanstd{9.478}{0.020} & \meanstd{18.5}{0.4} & \meanstd{8.950}{0.167} & \tablebest{\meanstd{3.5}{1.8}} & \meanstd{12.0}{2.6} & \meanstd{2.59}{0.06} & \meanstd{70.5}{5.9} & \meanstd{7.5}{1.5} \\
SA3 & \trainSR{} & \meanstd{9.478}{0.019} & \meanstd{18.5}{0.4} & \meanstd{8.942}{0.174} & \tablebest{\meanstd{3.5}{2.1}} & \meanstd{11.8}{2.7} & \meanstd{2.59}{0.06} & \meanstd{70.4}{6.1} & \meanstd{7.6}{1.3} \\
    \addlinespace[0.6pt]
SA4 & S & \meanstd{9.474}{0.034} & \meanstd{18.8}{0.7} & \meanstd{9.037}{0.281} & \tablethird{\meanstd{4.3}{3.7}} & \meanstd{13.6}{6.3} & \meanstd{2.60}{0.11} & \meanstd{72.6}{11.0} & \meanstd{7.9}{3.0} \\
SA4 & \trainSR{} & \meanstd{9.475}{0.033} & \meanstd{18.8}{0.7} & \meanstd{9.012}{0.283} & \tablethird{\meanstd{4.3}{3.8}} & \meanstd{13.2}{6.4} & \meanstd{2.60}{0.12} & \meanstd{72.1}{10.8} & \meanstd{7.9}{2.8} \\
    \bottomrule
  \end{tabular*}
  \par\vspace{3pt}
  \makebox[\textwidth][l]{\textbf{(b) Execution time and Real-based posterior-quality index}}
  \par\vspace{2pt}\noindent
  \begin{minipage}[t]{0.32\textwidth}
    \vspace{0pt}
    \begin{tabular*}{\linewidth}{@{\extracolsep{\fill}}lrr@{}}
      \toprule
      Model & Time [ms]$\downarrow$ & $I_{\mathrm{post}}\downarrow$ \\
      \midrule
      NP1 & \tablebest{\meanstd{0.35}{0.00}} & 6.50 \\
      NP2 & \tablesecond{\meanstd{0.45}{0.00}} & 4.67 \\
      NP3 & \meanstd{3.96}{0.00} & \tablethird{4.33} \\
      NP4 & \meanstd{1.49}{0.01} & \tablesecond{4.00} \\
      \textbf{NP5$^\star$} & \meanstd{2.83}{0.01} & \textbf{\tablebest{3.00}} \\
      \bottomrule
    \end{tabular*}
  \end{minipage}%
  \hfill
  \begin{minipage}[t]{0.32\textwidth}
    \vspace{0pt}
    \begin{tabular*}{\linewidth}{@{\extracolsep{\fill}}lrr@{}}
      \toprule
      Model & Time [ms]$\downarrow$ & $I_{\mathrm{post}}\downarrow$ \\
      \midrule
      SM1 & \tablesecond{\meanstd{0.45}{0.00}} & 9.00 \\
      SM2 & \meanstd{1.05}{0.00} & 8.67 \\
      SM3 & \meanstd{2.38}{0.05} & 8.17 \\
      SM4 & \meanstd{2.39}{0.02} & 9.67 \\
      \multicolumn{3}{c}{\strut} \\
      \bottomrule
    \end{tabular*}
  \end{minipage}%
  \hfill
  \begin{minipage}[t]{0.32\textwidth}
    \vspace{0pt}
    \begin{tabular*}{\linewidth}{@{\extracolsep{\fill}}lrr@{}}
      \toprule
      Model & Time [ms]$\downarrow$ & $I_{\mathrm{post}}\downarrow$ \\
      \midrule
      SA1 & \meanstd{0.49}{0.00} & 8.83 \\
      SA2 & \meanstd{1.09}{0.01} & 11.00 \\
      SA3 & \meanstd{2.73}{0.05} & 5.67 \\
      SA4 & \meanstd{2.75}{0.05} & 7.50 \\
      \multicolumn{3}{c}{\strut} \\
      \bottomrule
    \end{tabular*}
  \end{minipage}%
  \par\vspace{3pt}
  \parbox{\textwidth}{\footnotesize
  Mean $\pm$ sample standard deviation: five seeds for balanced performance (weighting:
  Section~\ref{sec:results-quantitative}); ten models, five per regime, for time.
  \tablebest{Blue}/\tablesecond{red}/\tablethird{green}: displayed-mean ranks 1/2/3, ties shared,
  across all rows per column in (a) and all 13 designs in (b). NP5$^\star$: MAGNETAR.
  NLL: target-weighted joint mass cross-entropy. Highest posterior density (HPD) gap:
  mean absolute empirical--nominal coverage error at masses $0.1,\ldots,0.9$, in percentage points (pp).
  Coverage counts soft-truth mass; support is the fraction of valid joint-pose candidates, both with proportional boundary ties.
  $V@40\%$: weighted mean HPD support interpolated at 40\% empirical coverage per seed.
  XY and circular-heading errors share the joint-MAP voxel; joint hits require errors within 1~m and $20^\circ$.
  Time: synchronized batch-one 32-bit floating-point (FP32) forward wall time to logits on NVIDIA H100 NVL,
  $18\!\times\!33\!\times\!33$ support.
  Index: average S and \trainSR{} Real metrics equally, then rank 13 designs (1 = best).
  Ascending $r_{\mathrm{NLL}},r_{\mathrm{XY}},r_\phi,r_{\mathrm{cal}},r_V$ rank
  NLL, MAP XY error, MAP heading error, HPD gap, and $V@40\%$, respectively.
  With $r_{\mathrm{pose}}=(r_{\mathrm{XY}}+r_\phi)/2$, $r_{\mathrm{unc}}=(r_{\mathrm{cal}}+r_V)/2$,
  $I_{\mathrm{post}}=(r_{\mathrm{NLL}}+r_{\mathrm{pose}}+r_{\mathrm{unc}})/3$.}
\end{table*}

\subsection{Multi-Seed Sim and Real Evaluation}
\label{sec:results-quantitative}

Table~\ref{tab:pose-global-probabilistic-five-seed}(a) assesses SP quality and
derived MAP decisions. Sim scores balance no-board/board conditions and
room groups within each; Real scores and Fig.~\ref{fig:pose-cone-benchmark}
also balance board placements, protocols, and RX sites hierarchically.

On Sim with independently sampled headings, all our NP variants improve NLL
over SM/SA; U-Net variants NP3--NP5 lower it by 0.8--0.9~nat relative to
the best parametric benchmark (uniform: 9.883~nat).
NP4 attains the lowest Sim NLL, with MAGNETAR within 0.06~nat. For MAGNETAR,
adding \realNB{} largely preserves held-out Sim performance: NLL rises by only
$0.013$~nat and highest posterior density (HPD) support at 40\% empirical
coverage, $V@40\%$, from 10.8\% to 10.9\%, while Real NLL falls by
$0.428$~nat.

Within the NP family on Real, adding \realNB{} lowers MAP errors for NP3/NP4
but worsens their NLL, whereas MAGNETAR improves both and attains the lowest
NLL and $V@40\%$ and highest joint hit rate, with $3.22$ times the
geometric-mean truth mass of the best \trainSR{} parametric baseline.
Conditioning the encoder on heading, rather than NP4's late query, thus
transfers best; NP4 retains lower individual MAP errors, but NP5 more often
satisfies both pose tolerances together.

\begingroup
\clubpenalty=10000\widowpenalty=10000
For true RX-to-TX bearing $\beta_0$, Fig.~\ref{fig:pose-cone-benchmark}
fixes the XY tolerance at 1~m and varies full opening $\alpha$, requiring
${|\wrap(\beta_{ij}-\beta_0)|}\leq\alpha/2$ and
${|\wrap(\phi^t_d-\phi^t_0)|}\leq\alpha/2$.
At fixed $\alpha$, the 1~m disk subtends a wider angle at shorter TX--RX
distances, so the bearing cone can further restrict XY acceptance;
farther away, the disk can lie within the cone.
Cone mass sums posterior probability in this region; hit rate tests the
global joint MAP. NP5 leads \trainSR{} MAP hit at every tested opening.
Normalized area under the curve (AUC) over $20^\circ$--$180^\circ$,
averaged equally over S and \trainSR{}, ranks NP5 first in MAP hit and third in mass.
\par
\endgroup

Table~\ref{tab:execution-posterior-index}(b) summarizes execution time and the
Real-based index $I_{\mathrm{post}}$, which ranks NP5 first overall, led by NLL
and $V@40\%$. NP5 executes faster than NP3 and slower than NP4.

\begin{figure}[!t]
  \centering
  \includegraphics[width=\fusionFigureWidth]{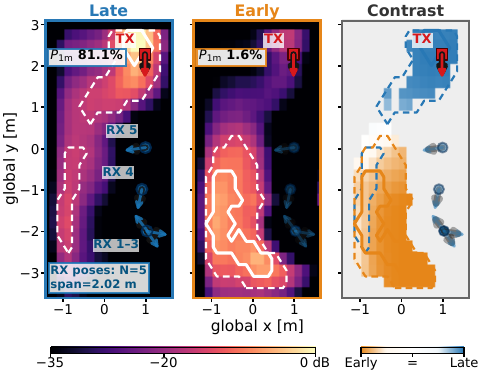}
  \caption{Measured-Real fusion illustration for $N=5$ \trainSR{} MAGNETAR
  (NP5) SPs under the shared-heading approximation. Solid and dashed contours
  mark 50\% and 90\% XY HPD mass, respectively. Contrast is
  \mbox{$(Q_{\mathrm{late}}-Q_{\mathrm{early}})/(Q_{\mathrm{late}}+Q_{\mathrm{early}})$}
  (Late: blue; Early: orange).}
  \label{fig:fusion-orders}
\end{figure}

\subsection{Enabling Heading-Consistent Fusion}
\label{sec:results-fusion}
\begingroup
\clubpenalty=10000\widowpenalty=10000

The joint-SP representation retains TX heading, enabling shared-heading fusion
before marginalization. Given a fixed TX pose, layout, and RX poses,
assume $N$ conditionally independent observations and a uniform prior on
common candidates. Let $\widehat p_n(d,i,j)$ be observation $n$'s joint SP
and $\widehat p_n(d\mid i,j)$ its conditional heading distribution. Then
\par
\endgroup
\begingroup
\predisplaypenalty=0
\allowdisplaybreaks[4]
\begin{align}
  Q_{\mathrm{early}}(i,j)
    &\propto\textstyle\prod\nolimits_{n=1}^{N}\sum\nolimits_d\widehat p_n(d,i,j),\notag\\
  Q_{\mathrm{late}}(i,j)
    &\propto\textstyle\sum\nolimits_d\prod\nolimits_{n=1}^{N}\widehat p_n(d,i,j)\notag\\
    &\propto\textstyle Q_{\mathrm{early}}(i,j)\,
      \underbrace{\textstyle\sum\nolimits_d\prod\nolimits_{n=1}^{N}\widehat p_n(d\mid i,j)}_{
        \substack{\kappa(i,j):\ \text{common-heading}\\\text{compatibility}}}.
  \label{eq:fusion-orders}
\end{align}
\par
\endgroup

\begingroup
\leavevmode\clubpenalty=10000\widowpenalty=10000
Here $Q_{\mathrm{early}}$ and $Q_{\mathrm{late}}$ are normalized TX-position
distributions, marginalizing heading before and after fusion, respectively.
Early discards which heading supports each location, admitting all $D^N$
tuples; Late retains only \mbox{$d_1=\cdots=d_N$}.
The factor $\kappa$ provides an interpretable consistency check, downweighting
locations requiring conflicting headings across views. If $\kappa$ is a
positive constant over Early's support, normalization makes both orders identical.

Fig.~\ref{fig:fusion-orders} fuses five \trainSR{} MAGNETAR Real SPs on a
common room-frame grid with fixed TX position and a shared-heading
approximation. Late suppresses the lower-room alternative, placing 81.1\%
mass within 1~m of truth versus 1.6\% for Early.

We evaluate 1,000 Sim sets (400 no-board, 600 board), each sharing TX pose
and layout across five RX poses. Averaging $P_{1\mathrm m}$ over five
\trainSR{} MAGNETAR models per set, mean $P_{1\mathrm m}$ is 24.62\% for
Late versus 23.67\% for Early, with a paired difference of $+0.95\pm6.25$
percentage points (mean $\pm$ sample standard deviation across sets).
Gains of at least 5 points occur in 11.0\% of sets, versus losses of that
magnitude in 5.1\%.
\par
\endgroup

\section{Conclusion}
We presented \method{}, which infers an explicit joint posterior over TX
position and heading from a single asynchronous multipath snapshot, so that
observations can be fused under a shared TX heading. Grid-based joint
posteriors outperformed parametric ones in simulation with random TX headings,
while the heading-conditioned U-Net achieved the lowest Real NLL and highest
joint hit rate with 1,800 measured training records.
Fusing joint posteriors improved on fusing marginals, markedly in one measured
case and modestly on average in simulation. Limitations include a single
measurement site, planar poses, and measured data in which the TX faces the RX.
Future work will study when shared-heading fusion helps, evaluate SP-guided
closed-loop search, and extend testing to more sites, independently varied
TX headings, and 3D poses.

\end{document}